\documentclass{article}

\usepackage{microtype}
\usepackage{graphicx}
\usepackage{subfigure}
\usepackage{booktabs} 

\usepackage{hyperref}

\usepackage[accepted]{icml2024}

\usepackage{amsmath}
\usepackage{amssymb}
\usepackage{mathtools}
\usepackage{amsthm}
\usepackage{algorithmic}
\usepackage[capitalize,noabbrev]{cleveref}

\theoremstyle{plain}

\theoremstyle{definition}

\theoremstyle{remark}

\usepackage[textsize=tiny]{todonotes}

\icmltitlerunning{ROOT}

\begin{document}

\twocolumn[
\icmltitle{ROOT: Robust Orthogonalized Optimizer for Neural Network Training}


\icmlsetsymbol{equal}{*}

\begin{icmlauthorlist}
\icmlauthor{Wei He}{equal,yyy}
\icmlauthor{Kai Han}{equal,yyy}
\icmlauthor{Hang Zhou}{yyy}
\icmlauthor{Hanting Chen}{yyy}
\icmlauthor{Zhicheng Liu}{yyy}
\icmlauthor{Xinghao Chen}{yyy}
\icmlauthor{Yunhe Wang}{yyy}\\
\textrm{$^1$Huawei Noah's Ark Lab}\\
\textrm{\small\{hewei142, kai.han, zhouhang25, chenhanting, liuzhicheng15, xinghao.chen, yunhe.wang\}@huawei.com}
\end{icmlauthorlist}

\icmlaffiliation{yyy}{Huawei Noah's Ark Lab}


\icmlkeywords{Machine Learning, ICML}

\vskip 0.3in
]



\printAffiliationsAndNotice{\icmlEqualContribution} 

\begin{abstract}
The optimization of large language models (LLMs) remains a critical challenge, particularly as model scaling exacerbates sensitivity to algorithmic imprecision and training instability. The recent advances in optimizer improve convergence efficiency through momentum orthogonalization but suffers from two key robustness limitations: dimensional fragility in orthogonalization precision and vulnerability to outlier-induced noise. 
To address these robustness challenges, we introduce ROOT, a \textbf{R}obust \textbf{O}rthogonalized \textbf{O}p\textbf{T}imizer that enhances training stability through dual robustness mechanisms. First, we develop a dimension-robust orthogonalization scheme using adaptive Newton iterations with fine-grained coefficients tailored to specific matrix sizes, ensuring consistent precision across diverse architectural configurations. Second, we introduce an optimization-robust framework via proximal optimization that suppresses outlier noise while preserving meaningful gradient directions. 
Extensive experiments demonstrate that ROOT achieves significantly improved robustness, with faster convergence and superior final performance compared to both Muon and Adam-based optimizers, particularly in noisy and non-convex scenarios. Our work establishes a new paradigm for developing robust and precise optimizers capable of handling the complexities of modern large-scale model training. The code will be available at \url{https://github.com/huawei-noah/noah-research/tree/master/ROOT}.
\end{abstract}

\section{Introduction}
\label{sec:intro}

The escalating computational demands of pre-training Large Language Models (LLMs)~\cite{brown2020language,openai2023gpt4,chen2025pangu} have positioned the design of optimization algorithms as a critical research frontier. The choice of an optimizer profoundly influences not only the convergence speed and final performance but also the substantial economic cost of model development. While Stochastic Gradient Descent (SGD)~\cite{robbins1951stochastic} established the foundational paradigm, the need for greater efficiency and stability in navigating complex, high-dimensional loss landscapes has driven the development of adaptive methods. The overarching goal is to design optimizers that are computationally efficient, robust, and scalable, enabling faster training of increasingly powerful models.

Adam~\cite{kingma2014adam} and its variant AdamW~\cite{loshchilov2017decoupled} have long been the de facto standards for training deep learning models. By incorporating momentum and per-parameter adaptive learning rates, Adam often achieves faster convergence than SGD. However, as model sizes surge into the billions of parameters, inherent limitations of adaptive methods become increasingly apparent, particularly numerical instability in mixed-precision training. Recent optimizers, such as Muon~\cite{jordan2024muon,liu2025muon}, represent architectural shifts aimed at addressing these challenges. For instance, Muon views weight matrices as holistic entities rather than independent scalars, employing a Newton-Schulz iteration to approximate orthogonal factors from momentum matrices. Such approaches have demonstrated benefits in training speed and memory efficiency compared to AdamW.

Despite these advances, our analysis identifies two critical limitations prevalent among those commonly-used optimizers. First, methods relying on fixed-coefficient iterative approximations (\emph{e.g.}, Newton-Schulz~\cite{jordan2024muon}) often exhibit a substantial precision gap, particularly for certain matrix dimensions where approximation errors are markedly higher. This stems from a one-size-fits-all approach to numerical stabilization, which fails to adapt to the diverse spectral properties of weight matrices across different layers. Second, many adaptive optimizers show heightened sensitivity to outlier-induced gradient noise, which is a common phenomenon in large-scale training where anomalous samples produce gradient components with disproportionately large magnitudes. This can corrupt update directions and destabilize training.

To address these limitations, we introduce \textbf{ROOT} (\textbf{R}obust \textbf{O}rthogonalized \textbf{O}p\textbf{T}imizer), a method designed to systematically enhance robustness from two key perspectives. First, to achieve \textit{algorithmic robustness} against structural uncertainties, we replace the fixed-coefficient Newton-Schulz iteration with an adaptive scheme employing fine-grained, dimension-specific coefficients. This ensures high-precision orthogonalization across all layers, making the process robust to variations in matrix dimensions. Second, for \textit{optimization robustness} against data-level noise, we incorporate a proximal optimization term that suppresses outlier-induced gradient noise via soft-thresholding, thereby stabilizing training without compromising convergence. Together, these contributions form a unified approach that significantly improves the robustness and reliability of the optimization process.

The main contributions of this work are summarized as follows:
\begin{itemize}
    \item We identify two key robustness limitations in the orthogonalization-based optimizers: lack of algorithmic robustness due to imprecise orthogonalization across varying matrix dimensions, and lack of optimization robustness against gradient outliers.
    
    \item We introduce a novel \textit{algorithmically robust} orthogonalization scheme via an adaptive Newton-Schulz iteration with dimension-specific coefficients, significantly improving orthogonalization accuracy across diverse network architectures.
    
    \item We develop an \textit{optimizationally robust} training framework through proximal optimization with soft-thresholding, providing theoretical guarantees for outlier suppression and stable convergence.
    
    \item Extensive experiments on LLM pre-training and fine-tuning demonstrate that ROOT achieves superior performance and faster convergence compared to state-of-the-art optimizers, while maintaining computational efficiency, particularly in noisy and non-convex scenarios.
\end{itemize}

\section{Related Work}
\label{sec:related_work}
In this section, we provide a brief overview of related work in the field of optimizers, as well as recent advances.

\subsection{Matrix-Aware Optimizers}
The training of deep neural networks has historically relied on stochastic first-order methods. Standard optimizers, such as SGD with momentum \citep{polyak1964some} and AdamW \citep{kingma2014adam, loshchilov2017decoupled}, treat model parameters as independent vectors. While computationally efficient, these coordinate-wise updates overlook the rich structural correlations inherent in weight matrices. To address this, second-order methods like K-FAC \citep{martens2015optimizing} and Shampoo \citep{gupta2018shampoo, anil2020scalable} leverage Kronecker-factored preconditioners to capture parameter geometry. However, despite their theoretical convergence advantages, these methods often incur prohibitive computational and memory overheads in large-scale settings.

Bridging this gap, the Muon optimizer \citep{jordan2024muon} regulate update geometry through matrix orthogonalization rather than explicit curvature approximation. Specifically, Muon employs a Newton-Schulz iteration~\cite{bernstein2024old,higham2008functions,guo2006schur} to orthogonalize the momentum matrix, which theoretically corresponds to performing steepest descent under the spectral norm \citep{li2025note, kovalev2025understanding}. This approach effectively balances structural awareness with computational efficiency, achieving $O(N)$ complexity similar to first-order methods while promoting coherent updates across parameter matrices.

\subsection{Recent Advances in Muon Variants}
Following Muon's success, several variants have emerged to extend its capabilities across different dimensions:

\textbf{Efficiency and Scalability.} To mitigate communication overhead in distributed settings, Dion \citep{ahn2025dion} replaces the dense Newton-Schulz iteration with amortized power iteration. For computational efficiency, LiMuon \citep{huang2025limuon} leverages randomized SVD, while Drop-Muon \citep{gruntkowska2025drop} explores randomized layer subsampling to reduce the update frequency.

\textbf{Adaptivity and Precision.} While Muon lacks element-wise adaptivity, recent works attempt to reintegrate it. AdaGO \citep{zhang2025adagrad} combines orthogonal directions with AdaGrad-style step sizes. AdaMuon \citep{si2025adamuon} incorporates a second-moment estimator, employing sign-based orthogonalization to enforce stability. On the numerical front, CANS \citep{grishina2025accelerating} utilizes Chebyshev polynomials to accelerate the convergence of the orthogonalization process over spectral intervals. Complementing these spectral-focused approaches, our work investigates orthogonalization precision across varying matrix dimensions.

\section{Approach}
\label{sec:method}
This section first outlines the preliminaries and then introduces our method for robust optimization which is achieved via adaptive Newton iteration with fine-grained coefficients and outlier suppression.

\subsection{Preliminaries}

The orthogonalization-based optimizers, \emph{e.g.}, Muon~\cite{jordan2024muon}, address the optimization of neural network parameters that exhibit a matrix structure. During each iteration $t$, the algorithm updates the weight matrix $\theta_{t-1}$ using the current momentum $\mu$, learning rate $\eta_t$, and objective function $\mathcal{L}$. The optimization procedure is defined by the following steps:

\begin{equation}
\begin{aligned}
M_t &= \mu M_{t-1} + \nabla \mathcal{L}(\theta_{t-1}) \\
M'_t &= \text{Newton-Schulz}(M_t) \\
\theta_t &= \theta_{t-1} - \eta_t M'_t
\end{aligned}
\end{equation}

In this formulation, $M_t$ represents the gradient momentum at step $t$, initialized as a zero matrix when $t=0$. The key innovation lies in the application of a Newton-Schulz (NS) iterative method~\cite{bernstein2024old} to approximate the transformation $(M_t M_t^{T})^{-1/2} M_t$. Considering the singular value decomposition (SVD) of $M_t = U \Sigma V^{T}$, this transformation yields $U V^{T}$, effectively orthogonalizing the momentum matrix. This orthogonalization process promotes isomorphic update matrices, which encourages the network to explore diverse optimization directions rather than converging along a limited set of dominant pathways.

The iterative process begins by initializing $X_0 = M_t / \|M_t\|_F$. At each subsequent step $k$, the matrix $X_k$ is updated from $X_{k-1}$ according to the recurrence relation:
\begin{equation}
\begin{aligned}
X_k =& a X_{k-1} + b X_{k-1} (X_{k-1}^T X_{k-1}) \\
&+ c X_{k-1} (X_{k-1}^T X_{k-1})^2
\end{aligned}
\end{equation}
After $N$ iterations, the resulting matrix $X_N$ serves as the approximation. The coefficients $a$, $b$, and $c$ are carefully selected to ensure proper convergence of the iterative scheme. The Muon optimizer adopts the values $a = 3.4445$, $b = -4.7750$, and $c = 2.0315$, which were originally designed to accelerate convergence for matrices with small initial singular values in 5 steps.

\begin{algorithm}[H]
	\caption{Muon Optimizer~\cite{jordan2024muon}}
	\textbf{Require:} Learning rate $\eta$, momentum $\mu$\\\vspace{-1em}
	\begin{algorithmic}[1]
		\STATE Initialize $M_0 \leftarrow 0$
		\FOR {$t = 1, \dots$}
		\STATE Compute gradient $G_t \leftarrow \nabla_{\theta}\mathcal{L}_t(\theta_{t-1})$
		\STATE $M_t \leftarrow \mu M_{t-1} + G_t$
		\STATE $M'_t \leftarrow \text{NewtonSchulz5}(M_t)$
		\STATE Update parameters $\theta_t \leftarrow \theta_{t-1} - \eta M'_t$
		\ENDFOR
		\STATE \textbf{Return} $\theta_t$
	\end{algorithmic}
\end{algorithm}

\subsection{Adaptive Newton Iteration with Fine-grained Coefficients}
\subsubsection{Enhancing Robustness against Matrix Dimension Variations}

The Muon optimizer's primary innovation lies in its use of the Newton-Schulz iteration to approximate an orthogonal matrix from the momentum matrix $M_t$. While computationally efficient, we identify a critical \textit{robustness} limitation: the fixed-coefficient NS iteration exhibits significant sensitivity to matrix dimensions, leading to inconsistent orthogonalization quality across different layers of deep neural networks.

The core issue stems from the one-size-fits-all approach to orthogonalization coefficients. As demonstrated in Table~\ref{tab:results_t5}, the mean squared error (MSE) of the orthogonal approximation varies dramatically with matrix shape. Square matrices ($n = m$) consistently yield the highest MSE values—up to two orders of magnitude worse than highly non-square configurations. For instance, with $m=2048, n=2048$, the MSE drops from 0.0499 to 0.0352 when the coefficients are learned adaptively instead of being fixed. This dimensional sensitivity creates an inherent \textit{fragility} in the optimization process, as layers with different dimensions receive orthogonalization of varying quality, compromising the consistency and reliability of gradient updates.

\begin{table}[htbp]
\centering
\caption{Orthogonalization error reveals dimensional fragility of fixed-coefficient Newton-Schulz iteration.}
\label{tab:results_t5}
\setlength{\tabcolsep}{1.5pt}
\resizebox{\columnwidth}{!}{%
\tiny 
\begin{tabular}{c|ccccccc}
\toprule
$n$ & 2048 & 4096 & 8192 & 2048 & 2048 & 2048 & 2048 \\
$m$ & 2048 & 4096 & 8192 & 3072 & 4096 & 8192 & 16384 \\
\midrule
$a$ & 3.4445 & 3.4445 & 3.4445 & 3.4445 & 3.4445 & 3.4445 & 3.4445 \\
$b$ & -4.7750 & -4.7750 & -4.7750 & -4.7750 & -4.7750 & -4.7750 & -4.7750 \\
$c$ & 2.0315 & 2.0315 & 2.0315 & 2.0315 & 2.0315 & 2.0315 & 2.0315 \\
MSE & 0.0499 & 0.0637 & 0.0761 & 0.0338 & 0.0362 & 0.0340 & 0.0425 \\
\midrule
$a$ & 3.3334 & 3.3732 & 3.3886 & 2.9091 & 2.7739 & 2.5925 & 2.7045 \\
$b$ & -4.2591 & -4.6134 & -4.9026 & -3.8108 & -3.4911 & -3.0010 & -3.2335 \\
$c$ & 1.7791 & 2.0576 & 2.3105 & 1.8600 & 1.6992 & 1.4138 & 1.5336 \\
MSE & 0.0352 & 0.0470 & 0.0587 & 0.0024 & 0.0010 & 0.0003 & 0.0003 \\
\bottomrule
\end{tabular}%
}
\end{table}

\begin{figure*}[h]
    \centering
    \includegraphics[width=0.9\textwidth]{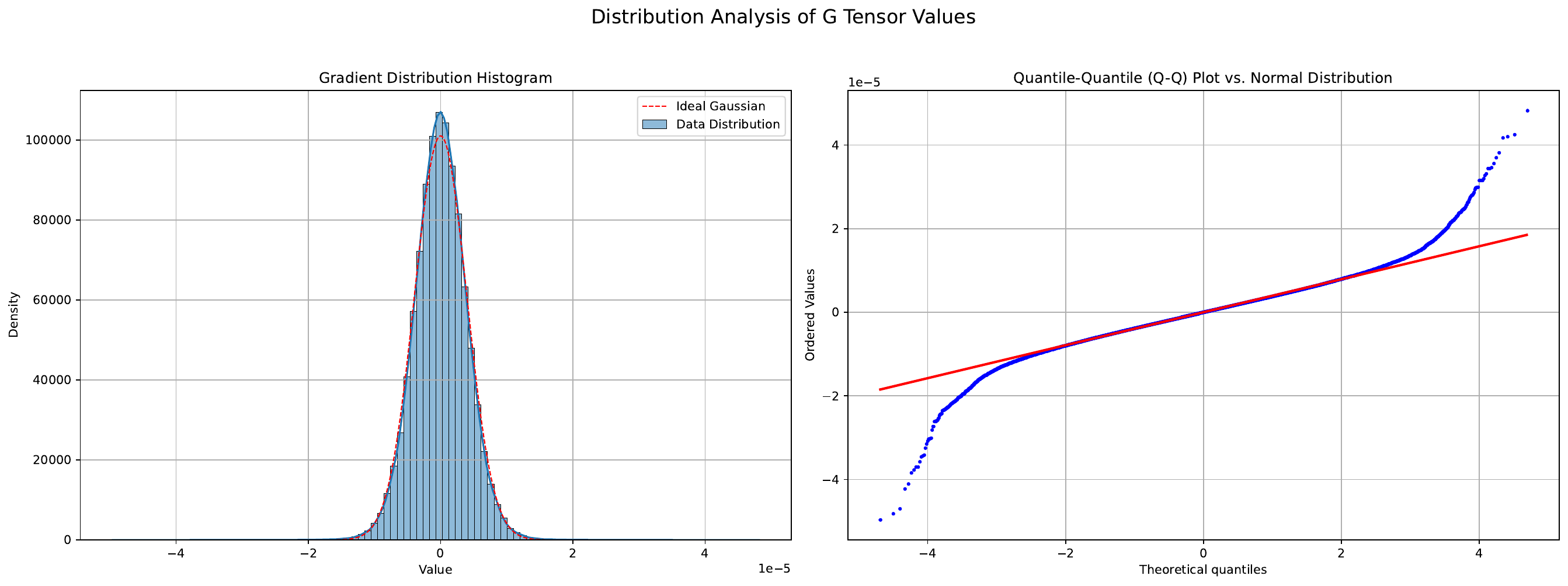}
    \caption{Analysis of gradient distribution revealing outlier characteristics. 
    \textbf{(Left)} Histogram with Gaussian reference shows long-tailed distribution. 
    \textbf{(Right)} Q-Q plot quantifies deviation from normality, where points deviating from the diagonal indicate outliers. 
    These outliers can disproportionately influence the optimization process.}
    \label{fig:gradient_outliers}
\end{figure*}

To address this dimensional fragility and build a \textit{dimension-robust} orthogonalization process, we propose an adaptive Newton-Schulz iteration (AdaNewton) with fine-grained, dimension-specific coefficients. Instead of using global constants that are optimal only for an ``average'' matrix shape, we learn specialized coefficients $\{a^{(m,n)}, b^{(m,n)}, c^{(m,n)}\}$ for each unique matrix size $(m, n)$ in the network architecture. This approach ensures consistent high-precision orthogonalization regardless of layer dimensions, making the optimization process robust to the inherent architectural variations in modern neural networks.

Formally, for a matrix $X_{k-1} \in \mathbb{R}^{m \times n}$, our robust adaptive update rule becomes:

\begin{equation}
\begin{aligned}
\label{eq:adaptive_update}
X_k =& a^{(m,n)} X_{k-1} + b^{(m,n)}  X_{k-1}( X_{k-1}^TX_{k-1}) \\
&+ c^{(m,n)} X_{k-1}( X_{k-1}^TX_{k-1})^2 
\end{aligned}
\end{equation}

The coefficients can be optimized jointly with the model parameters during training, allowing the orthogonalization process to automatically adapt to the specific spectral properties of each layer type. This fine-grained adaptation represents a paradigm shift from fragile dimension-sensitive orthogonalization to robust dimension-invariant orthogonalization, ensuring consistent update quality throughout the network.

\subsubsection{Theoretical Guarantees for Robust Convergence}

To establish the theoretical superiority of our adaptive coefficient design, we analyze the convergence properties of the Newton-Schulz iteration from a robust optimization perspective. The core insight is that by tailoring coefficients to specific matrix dimensions, we achieve a more precise approximation of the desired orthogonal transformation.

Consider the Newton-Schulz iteration defined by the polynomial mapping:
\[
g(x) = a x + b x^3 + c x^5
\]
After $T$ iterations, the cumulative transformation is given by the $T$-fold composition $g^{(T)}(x) = g(g(\cdots g(x)\cdots))$.

Now, assuming that $X_t$ can be decomposed via SVD as $X_t = U \Sigma_t V^T$, substituting into the iteration yields:
\[
X_{t+1} = U (a \Sigma_{t} + b \Sigma_{t}^3 + c \Sigma_{t}^5) V^T
\]
Thus, the iteration effectively operates on the singular values through the polynomial $g(x)$. After $T$ iterations, we have $\Sigma_T = g^{(T)}(\Sigma_0)$, and the optimization objective is to minimize the Frobenius norm distance to the identity matrix:
\begin{equation}\label{newton-loss}
\mathcal{L}_{newton}(a,b,c) = \|g^{(T)}(\Sigma) - I\|_F^2
\end{equation}

The standard fixed-coefficient approach selects parameters $(a,b,c)$ that minimize the worst-case error over a global singular value interval $I_{\text{std}} = [\sigma_{\min}, \sigma_{\max}]$:
\[
(a^*, b^*, c^*) = \arg\min_{a,b,c} \max_{\sigma \in I_{\text{std}}} |g^{(T)}(\sigma) - 1|
\]
This yields a convergence guarantee with error bound $E_{\text{std}}$.

However, this one-size-fits-all approach is inherently fragile because real-world weight matrices of different dimensions exhibit distinct singular value distributions $S^{(m,n)} \subseteq I_{\text{std}}$. Our robust adaptive method learns dimension-specific coefficients that minimize the error over these characteristic distributions:
\[
(a^{(m,n)}, b^{(m,n)}, c^{(m,n)}) = \arg\min_{a,b,c} \max_{\sigma \in S^{(m,n)}} |g^{(T)}(\sigma) - 1|
\]

The key theoretical advantage emerges from the optimization structure. Since $S^{(m,n)} \subseteq I_{\text{std}}$, the minimax error over the smaller set is bounded by the global error:
\begin{align*}
E_{(m,n)} &= \min_{a,b,c} \max_{\sigma \in S^{(m,n)}} |g^{(T)}(\sigma) - 1| \\
&\leq \min_{a,b,c} \max_{\sigma \in I_{\text{std}}} |g^{(T)}(\sigma) - 1| = E_{\text{std}}
\end{align*}

Moreover, when $S^{(m,n)}$ is a proper subset of $I_{\text{std}}$ (which occurs for non-square matrices), the adaptive coefficients can achieve strictly better approximation. Formally, if $S^{(m,n)} \subset I_{\text{std}}$ and the function $g^{(T)}(x)$ is not constant, then typically:
\[
E_{(m,n)} < E_{\text{std}}
\]

This improvement translates directly to our optimization objective $\|g^{(T)}(\Sigma) - I\|_F^2$. For a matrix with singular values $\{\sigma_i\}$, we have:
\[
\|g^T(\Sigma) - I\|_F^2 = \sum_i (g^{(T)}(\sigma_i) - 1)^2
\]
By achieving a smaller maximum error $\max_i |g^{(T)}(\sigma_i) - 1|$ through our adaptive coefficients, we obtain a tighter bound on the Frobenius norm objective in Eq.~\ref{newton-loss}. This theoretical result demonstrates that our dimension-adaptive approach provides provably better orthogonalization compared to the fixed-coefficient method, establishing the robustness advantages of our design.


\subsection{Robust Optimization via Outlier Suppression}
\subsubsection{Enhancing Robustness against Outlier-Induced Gradient Noise}
\label{sec:outlier}

In large-scale language model training, mini-batch gradients are frequently contaminated by outlier noise with gradient components of anomalously large magnitudes~\cite{zhang2025adagrad}. Figure~\ref{fig:gradient_outliers} provides a conceptual visualization of such outlier noise within a gradient distribution.

These outliers pose a critical threat to the stability of the orthogonalization process in Muon. The NS algorithm requires an initial normalization of the input matrix $M_t$ to ensure its singular values are within a specific range for convergence. The presence of extreme outlier noise in $M_t$ can distort this normalization step and, more critically, be amplified by the polynomial nature of the NS iteration. This amplification compromises the intended denoising effect of orthogonalization, as erratic directions associated with outliers are preserved in the parameter update, potentially destabilizing training and causing issues like exploding attention logits in Transformers.

To build an \textit{outlier-robust} optimization framework, we incorporate proximal optimization with soft-thresholding. This approach provides inherent protection against gradient contamination while maintaining the benefits of momentum orthogonalization.

\subsubsection{Soft-thresholding for Outlier Suppression}

Our robust optimization framework is derived from a gradient clipping perspective that explicitly handles outlier noise while preserving the overall gradient direction. We model the gradient (or momentum) $M_t$ as comprising two components: a base component $B_t$ containing the well-behaved gradient information, and an outlier component $O_t$ representing anomalous large-magnitude elements:

\begin{equation}
    M_t = B_t + O_t
    \label{eq:decomposition_model}
\end{equation}

The robust optimization objective aims to separate these components while constraining the influence of outliers:

\begin{equation}
    \min_{B_t, O_t} \|M_t - B_t - O_t\|_F^2 + \lambda\|O_t\|_1 \quad \text{subject to} \quad \|B_t\| \leq \tau
    \label{eq:clipping_objective}
\end{equation}

where $\lambda$ controls the sparsity of outlier detection and $\tau$ bounds the magnitude of the base component. This formulation explicitly penalizes the presence of large outlier components while ensuring the base component remains within reasonable bounds.

The optimal solution to this robust decomposition is characterized by the proximal operator for the L1-norm, which yields the soft-thresholding function~\cite{tibshirani1996regression,donoho2002noising}:
\begin{equation}
    \mathcal{T}_{\varepsilon}[x]_i = \text{sign}(x_i) \cdot \max(|x_i| - \varepsilon, 0)
    \label{eq:soft_threshold_general}
\end{equation}
where $\varepsilon$ is the threshold hyperparameter related to the regularization parameter $\lambda$. The soft-thresholding can be expressed as
\begin{equation}
\mathcal{T}_{\varepsilon}[x]_i =
\begin{cases}
x_i - \varepsilon, & \text{if } x_i > \varepsilon \\
0, & \text{if } |x_i| \leq \varepsilon \\
x_i + \varepsilon, & \text{if } x_i < -\varepsilon
\end{cases}
\end{equation}

This operation provides the closed-form solution for outlier separation:
\begin{equation}
\left\{
\begin{aligned}
O_t &= \mathcal{T}_{\varepsilon}(M_t), \quad~~ \textit{(sparse outlier components)} \\
B_t &= M_t - O_t. \quad \textit{(clipped, robust components)}
\end{aligned}
\right.
\label{eq:decomposition_solution}
\end{equation}

Mathematically, soft-thresholding can be interpreted as a continuous, differentiable alternative to hard clipping. While traditional gradient clipping abruptly truncates values beyond a fixed threshold, soft-thresholding applies a smooth shrinkage operation that preserves the relative ordering of gradient magnitudes while dampening extreme values.

\begin{algorithm}[H]
	\caption{ROOT Optimizer}
	\label{alg:robust_muon}
	\textbf{Require:} Learning rate $\eta$, momentum $\mu$, threshold $\varepsilon$\\\vspace{-1em}
	\begin{algorithmic}[1]
		\STATE Initialize $M_0 \leftarrow 0$
		\FOR {$t = 1, \dots$}
		\STATE Compute gradient $G_t \leftarrow \nabla_{\theta}\mathcal{L}(\theta_{t-1})$
		\STATE $M_t \leftarrow \mu M_{t-1} + G_t$ \quad \# Momentum accumulation
		\STATE $O_t \leftarrow \mathcal{T}_{\varepsilon}[M_t]$ \quad \# Outlier separation via soft-thresholding
		\STATE $B_t \leftarrow M_t - O_t$ \quad \# Clipped base components
		\STATE $B_t^{\text{orth}} \leftarrow \text{AdaNewton}(B_t)$ \quad \# Robust orthogonalization
		\STATE Update parameters $\theta_t \leftarrow \theta_{t-1} - \eta B_t^{\text{orth}}$
		\ENDFOR
		\STATE \textbf{Return} $\theta_t$
	\end{algorithmic}
\end{algorithm}

\begin{figure*}[t!]
    \centering
    \includegraphics[width=1\textwidth]{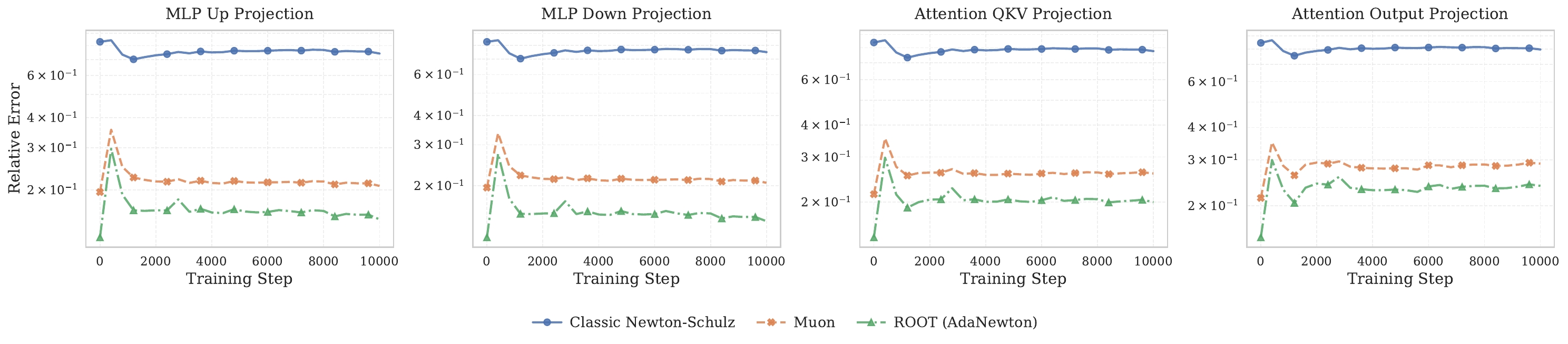}
    \vspace{-1em}
    \caption{\textbf{Orthogonalization precision relative to ground-truth SVD.} The plot tracks the Relative Error averaged over all optimized parameters (Attention QKV/O and MLP Up/Down projections). Under a fixed 5 iterations, ROOT maintains lower approximation error compared to the Muon baseline and Classic Newton-Schulz. This indicates that shape-specific coefficients provide superior fidelity across varying matrix dimensions.}
    \label{fig:decomp_metrics}
\end{figure*}

In the context of ROOT, we apply this robust decomposition to the momentum matrix $M_t \in \mathbb{R}^{m \times n}$:
\begin{equation}
    [\mathcal{T}_{\varepsilon}(M_t)]_{ij} = \mathcal{T}_{\varepsilon}([M_t]_{ij})
    \label{eq:matrix_threshold}
\end{equation}
The key innovation is that orthogonalization is applied only to the robust component $B_t$, while outlier components $O_t$ are discarded. This ensures that the orthogonalization process which is highly sensitive to large magnitude variations, operates on stable and clipped gradients, dramatically improving training robustness, as shown in Algorithm~\ref{alg:robust_muon}.

\section{Experiments}
\subsection{Implementation Details}
\paragraph{Experiments Setup}
To validate our approach, we conduct pretraining using the FineWeb-Edu dataset~\cite{penedo2024fineweb}, utilizing a 10-billion-token subset for ablation studies and a 100-billion-token sample for the main experiment. Both subsets are available at~\cite{huggingfacefw_2024}. To systematically evaluate the proposed optimization enhancements, we train a 1B Transformer~\cite{dubey2024llama,rang2025revealing}. All models are trained on distributed clusters of Ascend NPUs. The training infrastructure leverages high-speed interconnects to enable efficient data and model parallelism, ensuring scalable training across multiple nodes. An attention-mask-reset strategy~\cite{chen2025pangu} is also used to prevent self-attention between different documents within a sequence. 

Regarding the hyperparameters, all models are pretrained for a single epoch. We employ a cosine learning rate schedule that decays to $10\%$ of the peak learning-rate, following a warm-up phase of 2,000 steps. Specifically, for the 10B-token ablation experiments, the peak learning rate is set to $8 \times 10^{-4}$ with a global batch size of 0.4M. The pre-training sequence length is set to 4096. For the 100B-token main experiments, the learning rate is increased to $1.6 \times 10^{-3}$, and the global batch size is set to 1M. We adopt the default Muon hyperparameters from~\cite{jordan2024muon} and apply a 0.2 scaling factor following~\cite{liu2025muon} to align the update RMS with AdamW.

\paragraph{Evaluation Benchmark Setup} 
We evaluate our method on a comprehensive set of Academic benchmarks: HellaSwag~\cite{zellers-etal-2019-hellaswag}, ARC-easy (ARC-e) and ARC-challenge (ARC-c)~\cite{clark2018think}, BoolQ~\cite{clark-etal-2019-boolq}, PIQA~\cite{Bisk2020}, SciQ~\cite{Welbl2017CrowdsourcingMC}, WINO~\cite{sakaguchi2019winogrande}, OBQA~\cite{OpenBookQA2018}, and WSC~\cite{ea01b9c0db064caca6986b925d75f2bb}. 
We utilize the lm-evaluation-harness framework~\cite{eval-harness} for all evaluations, and we evaluate all tasks in zero-shot.

\subsection{Validation on Real Training Dynamics}
\label{sec:decomp_precision}

While Table~\ref{tab:results_t5} establishes the benefits of shape-specific coefficients on static distributions, we further validate whether these coefficients generalize to the dynamic spectral shifts observed during actual LLM training. We sampled input gradients $G_t$ from the first 10k pre-training steps, covering phases from early instability to stable convergence.

In this experiment, we compare three distinct orthogonalization strategies, each restricted to a strict budget of 5 iterations. 
We evaluate the Muon Baseline~\cite{jordan2024muon} with its general-purpose coefficients ($a=3.4445, b=-4.7750, c=2.0315$), as well as the Classic Newton-Schulz (Quintic) which uses coefficients ($1.875, -1.25, 0.375$) for order-5 convergence. 
We contrast these fixed strategies with our proposed ROOT (AdaNewton), which utilizes shape-specific coefficients learned from training gradients to explicitly target the geometry of different layers.

To quantify approximation fidelity, we report the Relative Error ($\| \hat{O} - O_{\text{SVD}} \|_F / \|O_{\text{SVD}}\|_F$) relative to the ground-truth SVD. 
The reported error is averaged across all parameter types targeted by the optimizer (Attention Query/Output and MLP Up/Down projections) at each step.

Figure~\ref{fig:decomp_metrics} illustrates the evolution of this metric. 
ROOT (Red) consistently maintains a lower relative error compared to the baselines throughout the training trajectory. 
While the Muon baseline (Blue) exhibits a distinguishable error floor, likely due to the sub-optimality of fixed coefficients for specific aspect ratios, ROOT achieves a closer approximation to the ground truth. 
These results suggest that incorporating dimension-aware coefficients mitigates the precision loss often associated with fixed-coefficient Newton-Schulz iterations on diverse matrix shapes.

\subsection{LLM Pre-training Analysis}
\label{sec:main_results}

\begin{figure}[H]
    \centering
    \includegraphics[width=1\linewidth]{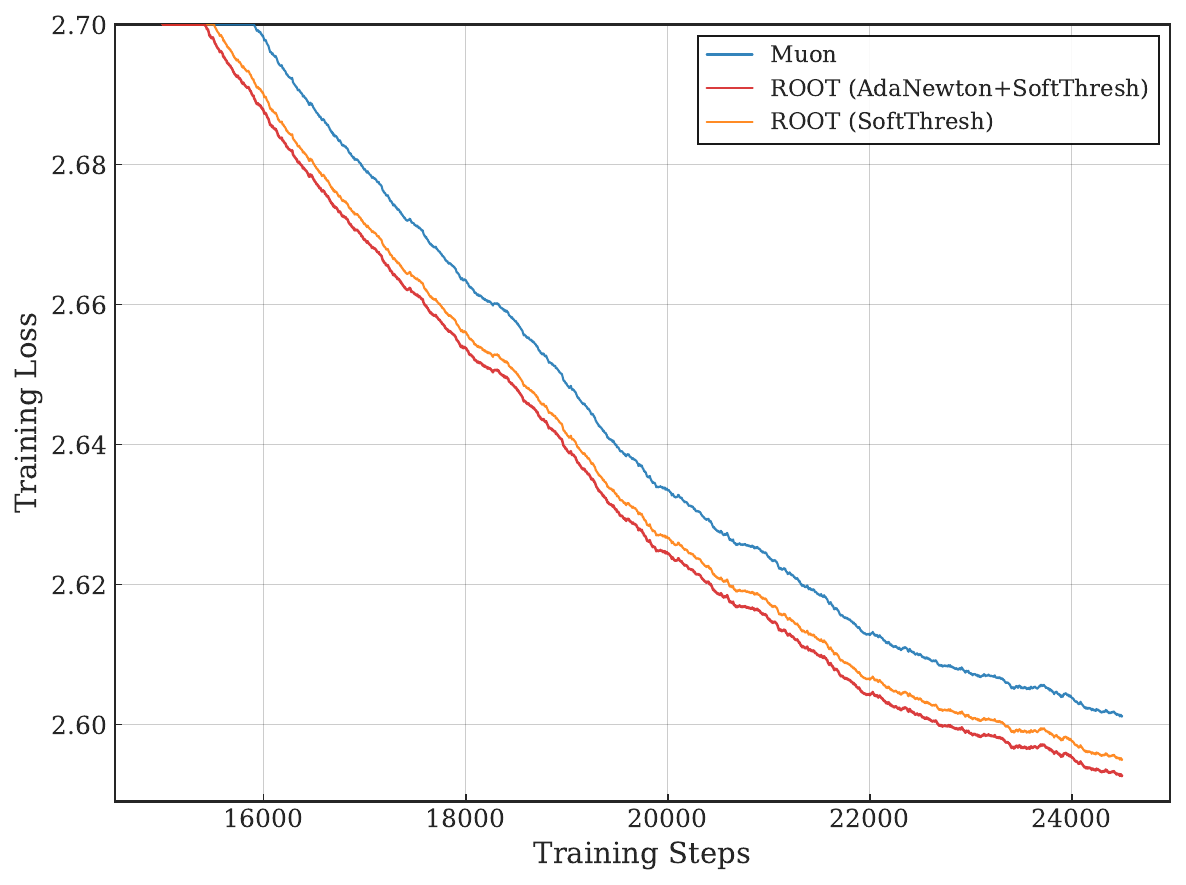}
    \caption{Training loss curves for 10B tokens. ROOT variants demonstrate faster convergence and lower final loss compared to Muon baseline, with full ROOT achieving the best performance.}
    \label{fig:pretraining_loss}
\end{figure}

We evaluate the pre-training performance of the proposed ROOT optimizer on the 1B Transformer over a 10B-token trajectory. Figure~\ref{fig:pretraining_loss} visualizes the training loss. We compare the Muon baseline against two configurations: \textit{ROOT-SoftThresh} (outlier suppression only) and the full ROOT optimizer.

As observed, both ROOT variants achieve consistently lower training loss compared to the Muon baseline. The ROOT (SoftThresh) variant improves convergence over standard Muon, suggesting that suppressing gradient outliers aids in stabilizing the optimization process. Furthermore, the full ROOT optimizer yields the lowest loss, indicating that dimension-adaptive orthogonalization provides additive benefits to outlier suppression. Ultimately, ROOT reaches a final training loss of 2.5407, surpassing the Muon baseline by 0.01.

\begin{table*}[t!]
\centering
\caption{Zero-shot performance on standard LLM benchmarks. ROOT outperforms both the AdamW baseline and the Muon optimizer across diverse common academic tasks.}
\label{tab:benchmark_results}
\vskip 0.15in
\begin{small}
\begin{sc}
\resizebox{\textwidth}{!}{
\begin{tabular}{lcccccccccc}
\toprule
Method & HellaSwag & BoolQ & PIQA & ARC-e & ARC-c & OBQA & SciQ & Wino & WSC & Avg. \\
\midrule
AdamW & 44.24 & \textbf{62.60} & 72.69 & 71.63 & \textbf{37.80} & 27.20 & 89.80 & 58.09 & 67.40 & 59.05 \\
Muon & 44.83 & 61.16 & 73.07 & \textbf{74.12} & 37.12 & 29.80 & 89.50 & 59.67 & 67.03 & 59.59 \\
ROOT & \textbf{45.37} & 62.08 & \textbf{73.12} & 72.14 & 36.86 & \textbf{31.20} & \textbf{90.40} & \textbf{60.30} & \textbf{69.60} & \textbf{60.12} \\
\bottomrule
\end{tabular}
}
\end{sc}
\end{small}
\vskip -0.1in
\end{table*}
\subsection{Benchmark Evaluation Results}
\label{sec:benchmark_results}

To assess model generalization, we further train the 1B model on 100B tokens and evaluate the trained 1B models across a diverse set of common academic benchmarks. Table~\ref{tab:benchmark_results} summarizes the zero-shot performance. ROOT achieves competitive or superior performance across the evaluated tasks. These results confirm that ROOT enhancements not only accelerate training convergence but also yield higher-quality final language models.

\subsection{Ablation Studies}
\label{sec:ablation}

\begin{figure}[h]
    \centering
    \includegraphics[width=1\linewidth]{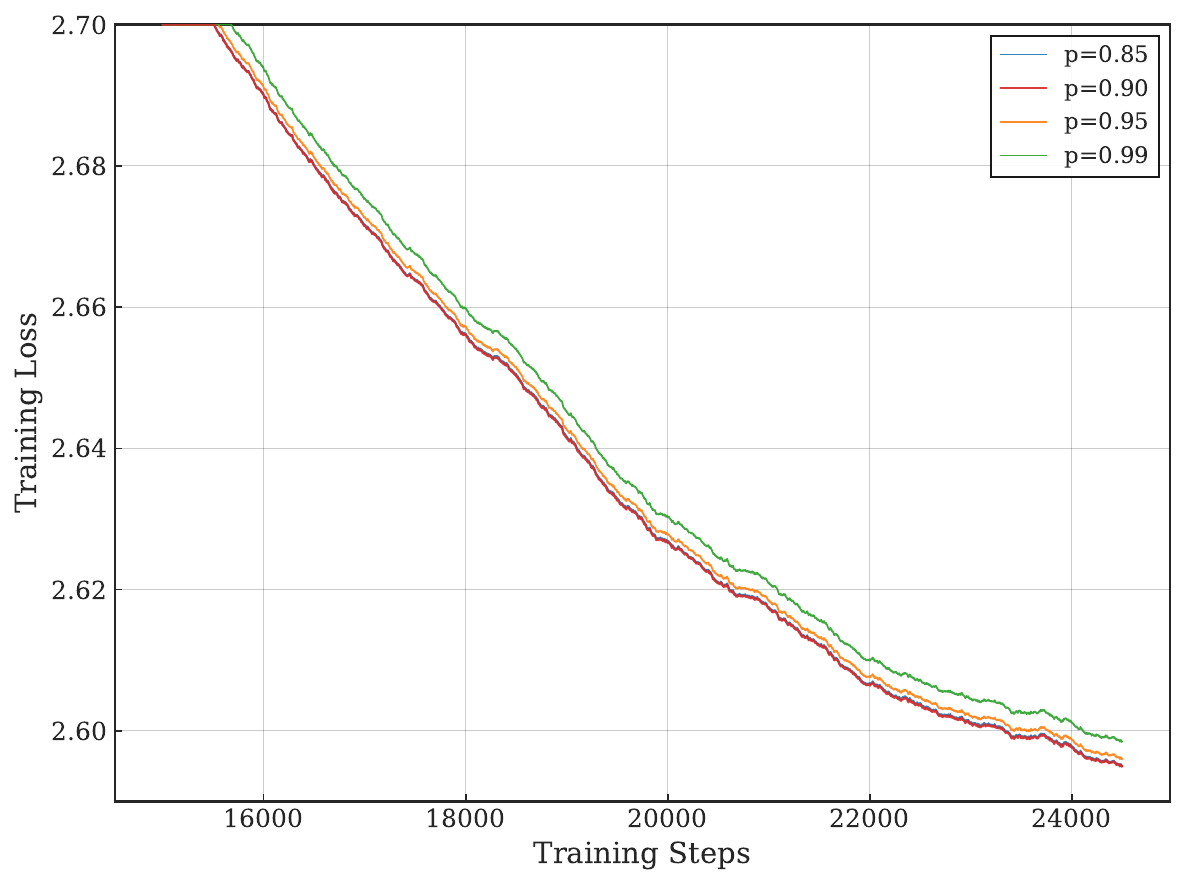}
    \caption{Ablation on the quantile hyperparameter $p$. The curve with $p=0.90$ demonstrates the optimal equilibrium between suppressing gradient noise and preserving informative gradient signals.}
    \label{fig:ablation_p}
\end{figure}

\paragraph{Outlier Suppression Threshold.}
Instead of a fixed threshold scalar, we employ a dynamic, percentile-based threshold $\varepsilon_t = \text{Quantile}(|M_t|, p)$ to adapt to the evolving gradient scale. We investigate the sensitivity of convergence to varying percentiles $p \in \{0.85, 0.90, 0.95, 0.99\}$.
As illustrated in Figure~\ref{fig:ablation_p}, the results reveal a trade-off between outlier suppression and signal preservation. A conservative threshold ($p=0.99$) results in under-suppression, failing to filter the heavy tail of the noise distribution and leading to sub-optimal stability. Conversely, an overly aggressive threshold ($p=0.85$) causes excessive truncation. We identify $p=0.90$ as the optimal equilibrium for LLM pre-training tasks, effectively isolating outliers while retaining the structural integrity of the gradients.

\paragraph{Spectral Calibration Strategy.}
To determine the optimal coefficients $\{a, b, c\}$ for ROOT (AdaNewton), we perform offline optimization using singular value distributions of momentum matrices with varying dimensions processed by the Muon optimizer, collected during the training process. We evaluate three calibration Strategies: Random Calibration, optimized solely on random Gaussian matrices; Mixed Calibration (1:1), where real \texttt{ns\_input} samples are augmented with random matrices at an equal ratio; and Mixed Calibration (1:3), which increases the proportion of random matrices to a 3:1 ratio.

\begin{figure}[h]
    \centering
    \includegraphics[width=1\linewidth]{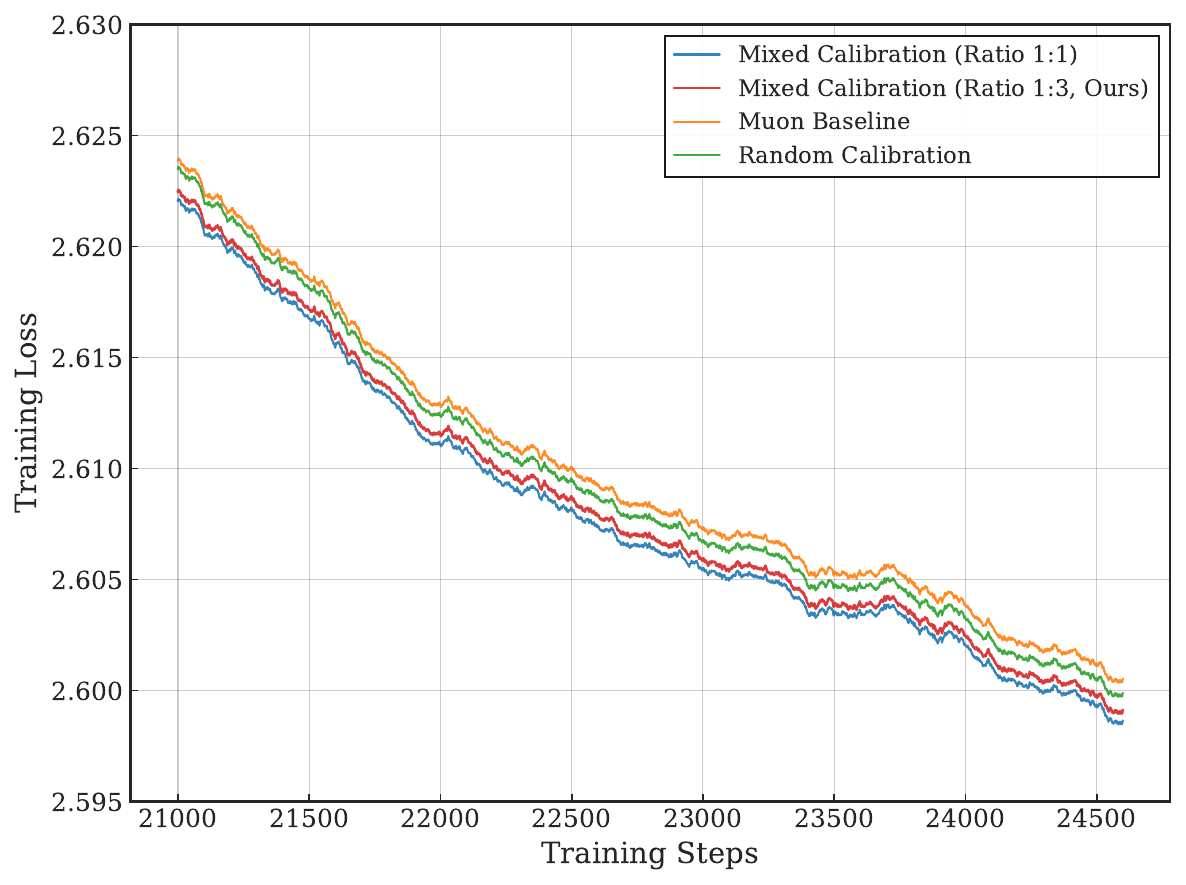}
    \caption{Ablation on the data composition for coefficient calibration. While a high ratio of real samples (Ratio 1:1) achieves lower loss here, it induces instability (loss spikes) in larger-scale experiments or when combined with ROOT (SoftThresh). The Mixed (1:3) strategy provides the optimal balance between convergence speed and robustness.}
    \label{fig:ablation_abc}
\end{figure}

Figure~\ref{fig:ablation_abc} illustrates the training trajectories under these configurations. While the Mixed (1:1) strategy yields the lowest loss in this specific setting, broader evaluations reveal its vulnerability to instability (e.g., loss spikes) when scaled to larger models or integrated with the Soft-Thresholding mechanism. Conversely, the Random Calibration baseline offers limited convergence acceleration. Consequently, we adopt the Mixed (1:3) strategy, which effectively prevents overfitting, thereby balancing accelerated convergence with generalization stability.

\subsection{Generalization to Vision Tasks}
\label{sec:vision_results}

To evaluate the generalization capabilities of ROOT beyond language modeling, we conducted image classification experiments by training a lightweight Vision Transformer ($\approx$ 6.3M parameters) on CIFAR-10 from scratch. We adopt a compact architecture adapted from \cite{VitCifar2023}, which processes images via $4 \times 4$ patches. In this experiment, we explicitly isolate the efficacy of the Soft-Thresholding mechanism against the Muon baseline. Models were trained for 100 epochs. Following standard protocols, all 2D weight matrices were optimized via Muon or ROOT, while 1D parameters (biases, norms) and embeddings and class-tokens were optimized using AdamW. The results in Table~\ref{tab:cifar_results} show that ROOT consistently outperforms the baseline. The improvement is most significant for the quantile percentile hyperparameter of 0.85, where the baseline achieves only 84.67\% accuracy. These results confirm that the soft-thresholding mechanism mitigates gradient noise, thereby enhancing generalization even in non-language modalities.

\begin{table}[h]
\centering
\caption{Top-1 Test Accuracy on CIFAR-10 (ViT, 6.3\,\textsc{m}, trained from scratch). Fixed params: Muon LR $= 0.02$, AdamW LR $= 0.001$, WD $= 5\times 10^{-5}$.}
\label{tab:cifar_results}
\vskip 0.15in
\resizebox{0.7\columnwidth}{!}{
\begin{small}
\begin{sc}
\begin{tabular}{lcc}
\toprule
Method  & $p$ (Quantile) & Acc (\%) \\
\midrule
Muon & - & 84.67 \\
ROOT   & 0.95 & 85.75 \\
ROOT  & 0.90 & 86.58 \\
ROOT   & 0.85 & \textbf{88.44} \\
\bottomrule
\end{tabular}
\end{sc}
\end{small}
}
\vskip -0.1in
\end{table}

\section{Conclusion}
\label{sec:conclusion}
In this work, we have presented ROOT, a robust orthogonalized optimizer that addresses two critical limitations in modern large-scale language model training. By introducing dimension-robust orthogonalization through adaptive Newton iterations and optimization robustness via proximal outlier suppression, ROOT establishes a new paradigm for stable and efficient neural network optimization. Our extensive experimental validation demonstrates that the proposed method achieves superior performance in challenging noisy and non-convex scenarios, providing both theoretical guarantees and practical benefits. This work opens promising directions for developing robust optimization frameworks that can handle the increasing complexity and scale of future language models, potentially enabling more reliable and efficient training of next-generation AI systems.

\bibliography{example_paper}
\bibliographystyle{icml2024}

%
%

\end{document}